\documentclass{article}
\usepackage{spconf,amsmath,graphicx,hyperref}
\usepackage{cite}
\usepackage{amssymb,amsfonts}
\usepackage{algorithmic}
\usepackage{textcomp}
\usepackage{xcolor}
\usepackage{subfigure} 
\usepackage{upgreek}
\usepackage{amsmath}
\usepackage{tabularx}
\usepackage{multirow}
\usepackage{enumitem}

\title{Accent-Invariant Automatic Speech Recognition via Saliency-Driven Spectrogram Masking}
\name{Mohammad Hossein Sameti$^{1}$, Sepehr Harfi Moridani$^{1}$, Ali Zarean$^{2}$, Hossein Sameti$^{1}$}
\address{$^{1}$Department of Computer Engineering, Sharif University of Technology \\
         $^{2}$Department of Computer Engineering, University of Tehran}

\begin{document}
%
\maketitle
\begin{abstract}
Pre-trained transformer-based models have significantly advanced automatic speech recognition (ASR), yet they remain sensitive to accent and dialectal variations, resulting in elevated word error rates (WER) in linguistically diverse languages such as English and Persian. To address this challenge, we propose an accent-invariant ASR framework that integrates accent and dialect classification into the recognition pipeline. Our approach involves training a spectrogram-based classifier to capture accent-specific cues, masking the regions most influential to its predictions, and using the masked spectrograms for data augmentation. This enhances the robustness of ASR models against accent variability. We evaluate the method using both English and Persian speech. For Persian, we introduce a newly collected dataset spanning multiple regional accents, establishing the first systematic benchmark for accent variation in Persian ASR that fills a critical gap in multilingual speech research and provides a foundation for future studies on low-resource, linguistically diverse languages. Experimental results with the Whisper model demonstrate that our masking and augmentation strategy yields substantial WER reductions in both English and Persian settings, confirming the effectiveness of the approach. This research advances the development of multilingual ASR systems that are resilient to accent and dialect diversity. Code and dataset are publicly available at: \href{https://github.com/MH-Sameti/Accent_invariant_ASR}{https://github.com/MH-Sameti/Accent\_invariant\_ASR}
\end{abstract}
\label{sec:intro}
\section{Introduction}

Automatic Speech Recognition (ASR) systems have evolved from providing transcription services for virtual assistants to enabling sophisticated healthcare applications~\cite{paraphasia}. This development demonstrates the critical role of ASR systems in enhancing accessibility and efficiency across various domains. Recent advancements in transformer-based models, such as the Whisper family, have significantly improved ASR performance by leveraging deep learning techniques to capture complex speech patterns~\cite{whisper}. These models have shown remarkable performance in transcribing spoken language across diverse contexts, including noisy environments and spontaneous conversations~\cite{ asradvancements}. However, despite their effectiveness, they indicate notable sensitivity to accent and dialect variations, particularly in linguistically diverse languages like English and Persian~\cite{racialdisp}. This sensitivity often results in high Word Error Rates (WER) when processing speech from speakers with non-native or regional accents, thereby limiting the accessibility and effectiveness of ASR technologies in global applications~\cite{improvedacc}.

Accents encapsulate unique phonetic and prosodic features that can obscure the underlying linguistic content, posing a substantial challenge for ASR systems trained mostly on standard or homogeneous datasets. These variations can lead to misinterpretations of phonemes and intonations, which are crucial for accurate speech recognition. Traditional approaches to mitigating accent-related discrepancies involve augmenting training datasets with diverse speech samples or fine-tuning models on accent-specific data~\cite{accentedspeech, improvedacc}. While these methods can improve performance, they often demand extensive data collection and may not generalize well to unseen accents or dialects, making them resource-intensive and less scalable.

Our main contributions are as follows:

\begin{itemize}[leftmargin=*]
  \item We propose a \textbf{saliency-driven spectrogram masking} framework that leverages Grad-CAM to identify accent-sensitive regions and suppress them, enabling ASR models to focus on accent-neutral linguistic features.
  
  \item We design a \textbf{lightweight, model-agnostic training strategy} that improves robustness to both known and unseen accents without requiring architectural modifications or full model retraining.
  
  \item We introduce the \textbf{P}ersian \textbf{D}ialect \textbf{ID}entification (PDID), a new multi-accent corpus covering 10 regional Persian accents, providing the first systematic benchmark for Persian accent robustness.
  
  \item We conduct extensive experiments on English (LibriSpeech, EdAcc, CommonAccent) and Persian (CommonVoice-fa, PDID), showing that our method consistently reduces WER/CER over SpecAugment baselines on accented speech~\cite{librispeech, edacc, commonaccent, commonvoice:2020}.
\end{itemize}


\section{Related Work}
Recent advances in transformer-based ASR models such as Whisper~\cite{whisper} have significantly improved speech recognition across noisy and spontaneous conditions. However, these models still exhibit notable sensitivity to accent and dialectal variations, with disproportionately high WER for non-native and regional speakers~\cite{racialdisp}.

More recently, large language model (LLM)-based approaches have been integrated into ASR pipelines to enhance robustness under accented and conversational speech~\cite{mu2024mmger, xu2025leveraging}.  
While these methods leverage powerful contextual reasoning to improve recognition, they drastically increase computational and memory costs, making them impractical for real-time or resource-constrained deployment.  
Moreover, their effectiveness diminishes in low-resource languages where training data and linguistic coverage are limited, reducing their utility for accent-heavy domains such as Persian.

A growing body of work focuses on enhancing accent robustness. Parameter-efficient adaptation methods like Mixture of Accent-Specific LoRAs (MAS-LoRA)~\cite{MixtureofLoRA} deploy accent-specialized LoRA experts, achieving improvements on accented corpora without full model retraining. Complementary to this, Qifusion-Net~\cite{Qifusion-Net} introduces a layer-adapted fusion strategy that dynamically integrates multi-accent acoustic features, reducing CER by over 20\% on large-scale benchmarks.

Beyond architecture, spectrogram manipulation and augmentation strategies remain underexplored for accent mitigation. While supervised contrastive learning has been applied to accented speech~\cite{han2021supervisedcontrastivelearningaccented}, direct masking of accent-related spectrogram regions has yet to be widely investigated—a gap our work explicitly targets to inject gradient information to the pipeline.


\section{Methodology}

\begin{figure}[t]
    \centering
    \includegraphics[width=0.9\linewidth]{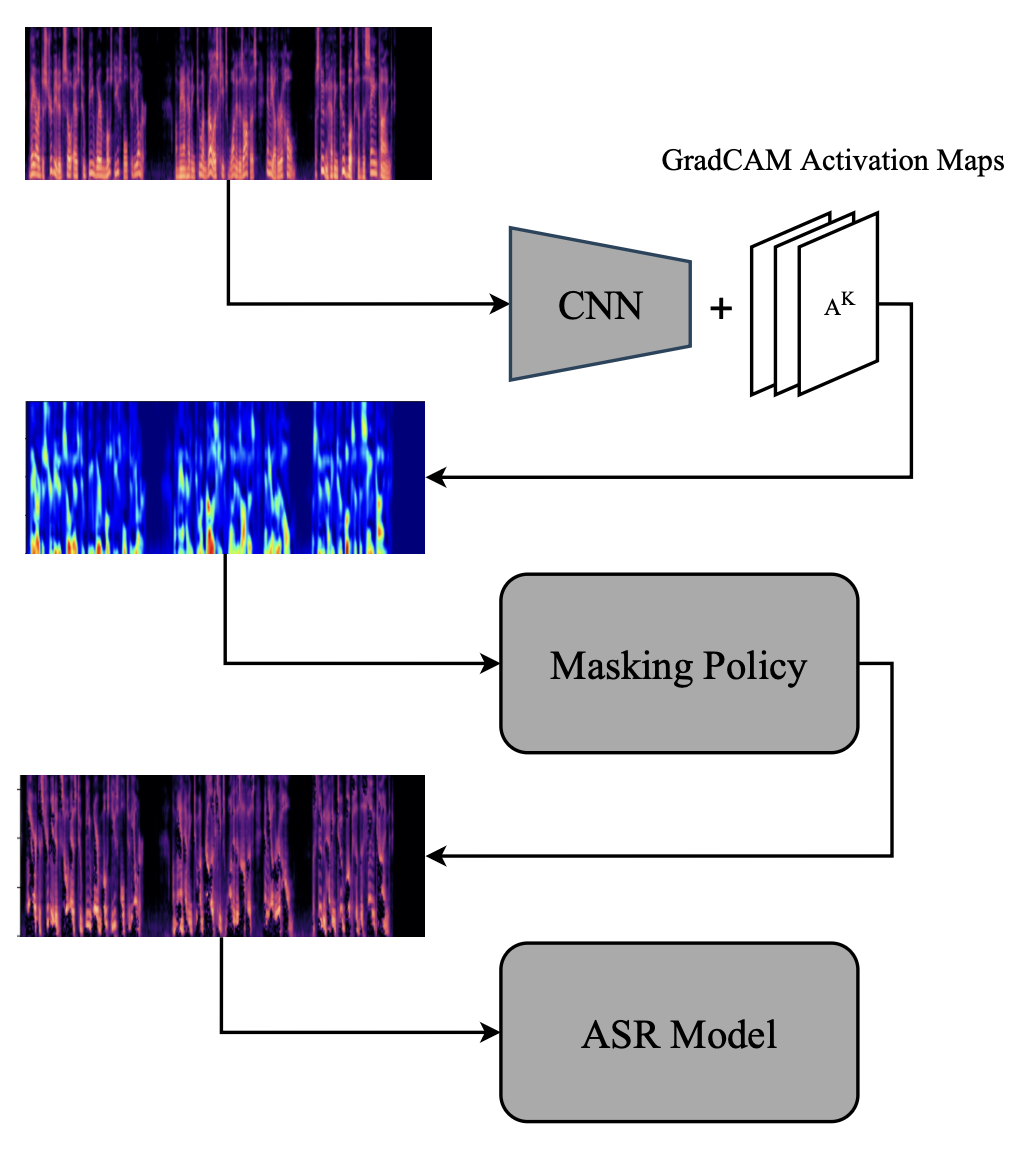}
    \caption{Overview of our accent suppression pipeline. First, the spectrogram is used to classify accents and generate a Grad-CAM saliency map highlighting accent-specific features. Next, a masking strategy is applied to suppress these accent-related regions while preserving essential information. Finally, the modified spectrogram is fed into the ASR model to improve generalization across diverse accents.}
    \label{fig:pipeline}
\end{figure}

This section details the proposed methodology for enhancing accent invariance in ASR systems. Our approach integrates accent and dialect classification into the ASR training pipeline through a multi-step process involving spectrogram-based classification, Grad-CAM for the localization of accent features, spectrogram masking, and fine-tuning a pre-trained ASR model on augmented data. The following subsections elaborate on each component of the method.

\subsection{PDID Dataset}
We collected speech samples from 10 regional Persian accents (Isfahani, Yazdi, Lori, Kurdish, Balochi, Southern, Northern, Tajiki, Mashhadi, and Shirazi) using sources such as local TV/radio and online platforms like Aparat and YouTube. Following a pipeline similar to the EMILIA dataset~\cite{he2024emilia}, we applied preprocessing steps including voice activity detection, speaker diarization, silence-based segmentation, and speech–music separation. All samples were standardized to 16kHz, mono-channel, 16-bit WAV format with normalized loudness, and segmented into 3–30 second clips. After quality filtering, about 23 hours of clean accent-labeled data remained from 200+ hours of raw speech, with Tajiki, Shirazi, and Balochi accents included only in the test set for robustness evaluation. Table~\ref{tab:sample_counts_pe} shows the distribution of samples and hours across the training accents.

\begin{table}[t]
    \centering
    \caption{Distribution of samples and hours across Persian accents in our dataset}
    \label{tab:sample_counts_pe}
    \begin{tabular}{l c c}
        \hline
        \textbf{Accent} & \textbf{Samples} & \textbf{Hours} \\
        \hline
        Isfahani & 996  & $\sim$2.2 h \\
        Yazdi    & 1114 & $\sim$2.4 h \\
        Shomali  & 7632 & $\sim$10.9 h \\
        Jonubi   & 147  & $\sim$1.1 h \\
        Lori     & 2220 & $\sim$4.0 h \\
        Kurdish  & 125  & $\sim$1.0 h \\
        Mashhadi & 379  & $\sim$1.5 h \\
        \hline
        \textbf{Train} & \textbf{12613} & \textbf{$\sim$23.0 h} \\  
        \hline
    \end{tabular}
\end{table}

\subsection{Accent Classification on Spectrograms}

To effectively identify accent-specific features within speech data, we first train an accent classifier using spectrogram representations of the input audio. Spectrograms provide a comprehensive visualization of the frequency content of speech signals over time, capturing both phonetic and prosodic characteristics essential for distinguishing accents.

We utilize a diverse dataset comprising speech samples from various accents and dialects of English. The resulting spectrograms are normalized to ensure consistent input scales for the classifier. Our accent classification dataset includes samples from the Edinburgh dataset for Southern British, Irish, Egyptian, and Italian and the LibriSpeech dataset for Standard English\cite{librispeech, edacc}. Table \ref{tab:sample_counts} contains the exact number of samples per accent.

\begin{table}[t]
    \centering
    \caption{Number of Samples per Class in the Dataset}
    \label{tab:sample_counts}
    \begin{tabular}{l c}
        \hline
        \textbf{Class} & \textbf{Number of Samples} \\
        \hline
        Standard & 1000 \\
        Southern British & 965 \\
        Irish & 704 \\
        Italian & 443 \\
        Egyptian & 346 \\
        Vietnamese & 332 \\
        \hline
        \textbf{Total} & \textbf{3790} \\
        \hline
    \end{tabular}
\end{table}

For accent classification, we utilize a convolutional neural network (CNN) architecture consisting of multiple convolutional layers with ReLU activations and max-pooling layers to capture hierarchical acoustic features. More specifically,  the architecture inputs normalized spectrograms of size $80 \times 3000$, where $80$ is the number of frequency bins and $3000$ is the number of time frames. Furthermore, four convolutional layers with 32, 64, 128, and 256 filters of size $3 \times 3$, each followed by ReLU activation. Max-pooling layers with a kernel size of $2 \times 2$ are applied after certain layers and dropout layers to prevent overfitting. Finally, a flattening layer is followed by a fully connected layer with 128 neurons and ReLU activation, including dropout for regularization, and a fully connected layer maps to the number of accent classes in the dataset.

The classifier is trained using the cross-entropy loss function and optimized with the Adam optimizer. Data augmentation techniques such as SpecAugment are applied during training to enhance the classifier's robustness to variability in speech signals\cite{specaugment}. The final accuracy of the classifier model is $74.6\%$ for English accents.  When applied with the same settings to our Persian accented dataset, 
the classifier achieved accuracy of $95\%$, 

\subsection{Masking Strategy}

To identify the regions in the spectrograms most indicative of accent-specific features, Gradient-weighted Class Activation Mapping (Grad-CAM) is utilized\cite{gradcam}, which provides a visual explanation by highlighting the areas of the input that significantly influence the classifier's decision.

Specifically, for each input spectrogram, the gradients of the predicted accent class are computed concerning the feature maps of the last convolutional layer. These gradients are then global-average-pooled to obtain weights, combined with the corresponding feature maps to produce a heatmap highlighting the salient regions associated with the accent classification.

A probabilistic masking strategy based on the normalized Grad-CAM scores is applied to suppress accent-specific features in the spectrograms. After normalizing the Grad-CAM activation map to obtain scores in the range $[0, 1]$, denoted as $C(i, j)$ for pixel $(i, j)$, a binary threshold mask $T(i, j)$ is defined as:
{\small
\begin{equation}
T(i, j) = 
\begin{cases}
1, & \text{if } C(i, j) > 0.3 \\
0, & \text{otherwise}.
\end{cases}
\end{equation}}

Next, random probability map $R(i, j)$ is generated, where each $R(i, j)$ is sampled from a uniform distribution over $[0, 1]$. Furthermore, $U(A, B)$ declares sampling from a random uniform distribution over $[A, B]$. The final mask $M(i, j)$ is computed as:
{\small
\begin{equation}
M(i, j) = 
\begin{cases}
1, & \text{if } T(i, j) = 0 \\
1, & \text{if } C(i, j) \geq 0.7 \text{ and } R(i, j) > 1 \\
1, & \text{if } 0.5 \leq C(i, j) < 0.7 \text{ and } R(i, j) > U(0.7, 0.9) \\
1, & \text{if } C(i, j) < 0.5 \text{ and } R(i, j) > U(0, 0.05) \\
0, & \text{otherwise}.
\end{cases}
\end{equation}}

\begin{table*}[t]
  \centering
  \caption{WER/CER results for English datasets (LibriSpeech, EdAcc, Unseen accents, and CommonAccent). 
  WsPr\_t: Whisper\_tiny, 
  WsPrLS\_t: WhisperLS\_tiny, 
  WsPrSAug\_t: SpecAugment baseline, 
  ARWsPr\_t: ours}
  \label{tab:results_en}
  \begin{tabular}{l|c|c|c|c}
    \hline
    \textbf{Model} & \textbf{LS} & \textbf{Accented} & \textbf{Unseen} & \textbf{CMA}\\
    \hline
    WsPr\_t~\cite{whisper}              & $8.0$ / $3.2$   & $42.0$ / $37.7$ & $34.7$ / $26.7$ &  $62.2$/$38.5$\\
    WsPrLS\_t~\cite{librispeech}             & $7.0$ / $2.7$   & $26.1$ / $16.0$ & $29.3$ / $19.4$  & $36.3$/$18.5$\\
    WsPrSAug\_t~\cite{specaugment}  & $7.3$ / $2.9$   & $27.0$ / $17.8$ & $30.1$ / $20.3$ & $38.3$/$21.6$  \\
    ARWsPr\_t (ours)           & \textbf{6.8} / \textbf{2.7} & \textbf{23.4} / \textbf{15.1} & \textbf{26.7} / \textbf{17.9} & \textbf{34.8}/\textbf{18.2}\\
    \hline
  \end{tabular}
\end{table*}

\begin{table*}[t]
  \centering
  \caption{WER/CER results for Persian datasets (CommonVoice-fa and regional accents). 
  WsPr\_b: Whisper\_base, 
  WsPrCV\_b/m: Whisper fine-tuned on CommonVoice (base/medium), 
  SpcAug: SpecAugment, 
  ARWsPr: ours , ARWsPr++: ours with GradCam++}
  \label{tab:results_fa}
  \begin{tabular}{l|c|c}
    \hline
    \textbf{Model} & \textbf{Standard} & \textbf{Accented} \\
    \hline
    WsPr\_b~\cite{whisper}            & $186.4$ / $209.4$ & $128.6$ / $93.6$ \\
    WsPrCV\_b~\cite{commonvoice:2020}           & $62.2$ / $25.4$   & $97.2$ / $61.7$ \\
    WsPrSpcAug\_b~\cite{specaugment}    & \textbf{61.5} / $23.9$   & $92.8$ / $51.6$ \\
    
    ARWsPr++\_b~\cite{Chattopadhay_2018}   & $62.4$ / $22.1$ & $90.3$ / $41.5$ \\
    ARWsPr\_b (ours) & $61.9$/ \textbf{21.1} & \textbf{88.8} / \textbf{40.6} \\
    \hline
    WsPr\_m~\cite{whisper} & $68.3$ / $32.1$ & $129.5$ /  $89.1$\\
    WsPrCV\_m~\cite{commonvoice:2020}           & \textbf{30.7} / \textbf{$8.9$}    & $70.4$ / $42.5$ \\
    ARWsPr\_m  (ours)          & $31.1$ / $9.9$ & \textbf{67.5} / \textbf{36.5} \\
    \hline
  \end{tabular}
\end{table*}

In this strategy:
\begin{itemize}
    \item If a pixel belongs to the region where $T(i,j)=0$, it always remains unchanged.
    \item If a pixel is located in a region that is considered strongly accent-related ($C(i, j) \geq 0.7$) all such pixels are masked.
    \item If a pixel belongs to the region with a moderate to high score ($ 0.5 \leq C(i, j) <  0.7$), it is masked with a probability between 0.7 and 0.9 using a uniform probability distribution ($U(0.7,0.9)$), This ensures that nonrelevant pixels have a chance to be included in accent-related features, thereby reducing errors to some extent.
    \item If a pixel falls within the low to moderate score ($C(i, j) < 0.5$), it is masked with a probability between $U(0,0.05)$, to account for accent-related regions that might have been mistakenly assigned a low score, thus mitigating errors to some extent.

\end{itemize}

 The masked spectrogram is then generated by element-wise multiplication of the original spectrogram with the mask $M(i, j)$:
{\small
\begin{equation}
\text{Masked Spectrogram}(i, j) = \text{Spectrogram}(i, j) \times M(i, j).
\end{equation}
}
This probabilistic masking strategy ensures that accent-related features are suppressed while retaining essential linguistic information, enhancing the ASR model's ability to generalize across different accents. As shown in Figure \ref{fig:pipeline}, the original spectrogram, Grad-CAM activation map, and masked spectrogram illustrate the accent feature localization and suppression process.

The masked spectrograms are combined with the primary dataset to form an augmented training dataset. This augmentation encourages the ASR model to learn accent-neutral representations by exposing it to accented and accent-suppressed versions of the same speech samples. Leveraging this dataset, we fine-tune a state-of-the-art transformer-based ASR model to improve its robustness in accent and dialect variations. 


\section{Experiments and Results}

We conducted experiments on both English and Persian datasets to evaluate the effectiveness of our proposed accent-aware masking method. As Table \ref{tab:results_en} shows For English, we used LibriSpeech, EdAcc, and  CommonAccent, while Table \ref{tab:results_fa} shows for Persian we used the CommonVoice (fa) subset along with our newly collected accented dataset PDID. Training was performed on NVIDIA RTX 3090 GPUs using AdamW optimizer, with learning rates of $1 \times 10^{-5}$ for tiny and $3 \times 10^{-6}$ for base/medium models, batch sizes of 32, 16, and 4 respectively, and 10 epochs. The evaluation metrics were WER and CER, complemented by ablations using Grad-CAM and Grad-CAM++ to generate accent-masking policies. Results show that our method significantly outperforms both pre-trained Whisper and LibriSpeech fine-tuned baselines, as well as a SpecAugment baseline, particularly in accented and unseen-accent settings~\cite{Chattopadhay_2018}. For Persian, fine-tuning on CommonVoice (fa) improves performance, but our accent-masked approach yields further gains across both base and medium sizes.

Overall, these results confirm that accent-masked training consistently reduces CER and WER across both English and Persian. The improvements are particularly strong on unseen accents, highlighting the robustness and generalizability of the proposed method.

\section{Conclusion}
We proposed a saliency-driven spectrogram masking framework that uses Grad-CAM to suppress accent-specific features and encourage ASR models to learn accent-neutral representations. Our approach is lightweight, model-agnostic, and improves robustness without architectural modifications or full retraining. In addition, we introduced the \textbf{PDID} dataset, the first multi-accent benchmark for Persian ASR covering 10 regional dialects. Experiments on English  and Persian showed consistent WER/CER reductions, with relative gains up to \textbf{14\%} on accented speech compared to SpecAugment baselines. These results confirm that targeted spectrogram masking is an effective strategy for accent-robust ASR.

\bibliographystyle{IEEEbib}
\bibliography{main}

\end{document}